\documentclass{acm_proc_article-sp}

\begin{document}
%
\CopyrightYear{2004} 

\title{Learning Optimal Augmented Bayes Networks}
%
%

\numberofauthors{2}
%

\author{
%
\alignauthor Vikas Hamine\\
       \affaddr{Dept. of Computer Science}\\
       \affaddr{University of New Mexico}\\
       \affaddr{Albuquerque, New Mexico 87131 USA}\\
       \email{vikas@cs.unm.edu}
\alignauthor Paul Helman\\
       \affaddr{Dept. of Computer Science}\\
       \affaddr{University of New Mexico}\\
       \affaddr{Albuquerque, New Mexico 87131 USA}\\
       \email{helman@cs.unm.edu}
}
\date{27 February 2004}
\maketitle
\begin{abstract}
Naive Bayes is a simple Bayesian classifier with strong independence assumptions among the attributes. This classifier, despite its strong independence assumptions, often performs well in practice. It is believed that relaxing the independence assumptions of a naive Bayes classifier may improve the classification accuracy of the resulting structure. While finding an 
optimal unconstrained Bayesian Network (for most any reasonable scoring measure) is an NP-hard problem, it is possible to learn in polynomial time optimal networks obeying various structural restrictions. Several authors have examined the possibilities of adding augmenting arcs between attributes of a Naive Bayes classifier. Friedman, Geiger and Goldszmidt define the TAN structure in which the augmenting arcs form a tree on the attributes, and present a polynomial time algorithm that learns an optimal TAN with respect to MDL score. Keogh and Pazzani define Augmented Bayes networks in which the augmenting arcs form a forest on the attributes, and present heuristic search methods for learning good, though not optimal, augmenting arc sets. In this paper, we present a simple, polynomial time greedy algorithm for learning an optimal Augmented Bayes Network with respect to MDL score.
\end{abstract}
\section{Introduction}
Classification is a machine learning task that requires construction of a function that classifies examples into one of a discrete set of possible categories. Formally, the examples are vectors of \emph{attribute} values and the discrete categories are the \emph{class} labels. The construction of the classifier function is done by training on preclassified instances of a set of attributes. This kind of learning is called \emph{supervised} learning as the learning is based on labeled data. A few of the various approaches for supervised learning are artificial neural networks, decision tree learning, support vector machines and Bayesian networks \cite{mitchell:machine}. All these methods are comparable in terms of classification accuracy. Bayesian networks are especially important because they provide us with useful information about the structure of the problem itself.

One highly simple and effective classifier is the naive Bayes classifier \cite{duda:pattern}. The naive Bayes classifier is based on the assumption that the attribute values are conditionally independent of each other given the class label. The classifier learns the probability of each attribute $X_i$ given the class $C$ from the preclassified instances. Classification is done by calculating the probability of the class $C$ given all attributes $X_1, X_2, ..., X_n$. The computation of this probability is made simple by application of Bayes rule and the rather naive assumption of attribute independence. In practical classification problems, we hardly come across a situation where the attributes are truly conditionally independent of each other. Yet the naive Bayes classifier performs well as compared to other state-of-art classifiers.

An obvious question that comes to mind is whether relaxing the attribute independence assumption of the naive Bayes classifier will help improve the classification accuracy of Bayesian classifiers. In general, learning a structure (with no structural restrictions) that represents the appropriate attribute dependencies is an NP-Hard problem. Several authors have examined the possibilities of adding arcs (augmenting arcs) between attributes of a naive Bayes classifier that obey certain structural restrictions. For instance, Friedman, Geiger and Goldszmidt \cite{friedman:bayesian} define the TAN structure in which the augmenting arcs form a tree on the attributes. They present a polynomial time algorithm that learns an optimal TAN with respect to MDL score. Keogh and Pazzani \cite{keogh:learning} define Augmented Bayes networks in which the augmenting arcs form a forest on the attributes (a collection of trees, hence a relaxation of the structural restriction of TAN), and present heuristic search methods for learning good, though not optimal, augmenting arc sets. The authors, however, evaluate the learned structure only in terms of observed misclassification error and not against a scoring metric, such as MDL. Sacha in his dissertation (unpublished, $http://jbnc.sourceforge.net/JP\_Sacha\_PhD\_Dissertat
i\\on.pdf$), defines the same problem as Forest Augmented Naive Bayes (FAN) and presents polynomial time algorithm for finding good classifiers with respect to various quality measures (not MDL). The author however, does not claim the learned structure to be optimal with respect to any quality measure.

In this paper, we present a polynomial time algorithm for finding optimal Augmented Bayes Networks/Forest Augmented Naive Bayes with respect to MDL score. The rest of the paper is organized as follows. In section 2, we define the Augmented Bayes structure. Section 3, defines the MDL score for Bayesian Networks. The reader is referred to the Friedman paper \cite{friedman:bayesian} for details on MDL score, as we present only the necessary details in section 3. Section 4 provides intuition about the problem and Section 5 and 6 present the polynomial time algorithm and prove that its optimal.
\section{Augmented Bayes Networks}
The Augmented Bayes Network (ABN) structure is defined by Keogh and Pazzani \cite{keogh:learning} as follows:
\begin{itemize}
\item Every attribute $X_i$ has the class attribute $C$ as its parent.
\item An attribute $X_i$ may have at most one other attribute as its parent.
\end{itemize}
Note that, the definition is similar to the TAN definition given in \cite{friedman:bayesian}. The difference is that whereas TAN necessarily adds $n-1$ augmenting arcs (where $n$ is the number of attributes); ABN adds any number of augmenting arcs up to $n-1$. Figure 1 shows a simple ABN. The dashed arcs represent augmenting arcs. Note that attributes $1$ and $5$ in the figure do not have any incoming augmenting arcs. Thus the ABN structure does not enforce the tree structure of TAN, giving more model flexibility.
\begin{figure}
\centering
\psfig{file=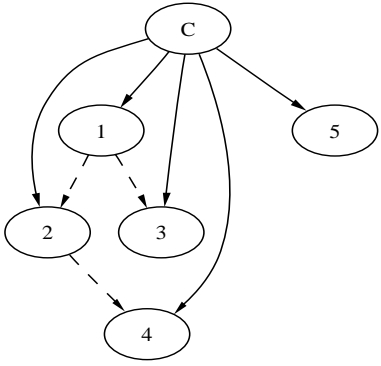, height=2in, width=2in}
\caption{A simple Augmented Bayes Network}
\end{figure}
\section{Background}
In this section we present the definitions of Bayesian network and its MDL score. This section is derived from the Friedman paper \cite{friedman:bayesian}. We refer the reader to the paper \cite{friedman:bayesian} for more information as we only present the necessary details.

A Bayesian network is an annotated directed acyclic graph (DAG) that encodes a joint probability distribution of a domain composed of a set of random variables (attributes). Let $U = \{X_{1},..., X_{n}\}$ be a set of $n$ discrete attributes where each attribute $X_{i}$ takes values from a finite domain. Then, the Bayesian network for $U$ is the pair $B = <G,\Theta>$, where $G$ is a DAG whose nodes correspond to the attributes $X_{1},..., X_{n}$ and whose arcs represent direct dependencies between the attributes. The graph structure $G$ encodes the following set of independence assumptions: each node $X_{i}$ is independent of its non-descendants given its parents in $G$. The second component of the pair $\Theta$ contains a parameter $\theta_{x_{i}|\Pi_{x_{i}}} = P(x_{i}|\Pi_{x_{i}})$ for each possible value $x_{i}$ of $X_{i}$ and $\Pi_{x_{i}}$ of $\Pi_{X_{i}}$. $B$ defines a unique joint probability distribution over $U$ defined by:
\[ P_{B}(X_{1},...,X_{n}) = \prod_{i = 1}^{n} P_{B}(X_{i}|\Pi_{X_{i}}) \]
The problem of learning a Bayesian network can be stated as follows. Given a \emph{training set} $D = \{u_{1},..., u_{N}\}$ of instances of $U$, find a network that best fits $D$.

We now review the \emph{Minimum Description Length} (MDL) \cite{rissanen:modelling} of a Bayesian Network. As mentioned before, our algorithm learns optimal ABNs with respect to MDL score. The MDL score casts learning in terms of data compression. The goal of the learner is to find a structure that facilitates the shortest description of the given data \cite{friedman:bayesian, friedman:discretization}. Intuitively, data having regularities can be described in a compressed form. In context of Bayesian network learning, we describe the data using DAGs that represent dependencies between attributes. A Bayesian network with the least MDL score (highly compressed) is said to model the underlying distribution in the best possible way. Thus the problem of learning Bayesian networks using MDL score becomes an optimization problem. The MDL score of a Bayesian network $B$ is defined as
\begin{equation}
MDL(B) = \frac{|B|\log{N}}{2} - N\sum_i^nI(X_i; \Pi_{X_i})
\end{equation}
where,
$N$ is the number of instances of the set of attributes, $|B|$ is number of parameters in the Bayesian network $B$, $n$ is number of attributes, and $I(X_i; \Pi_{X_i})$ is the mutual information between an attribute $X_i$ and its parents in the network. As per the definition of the ABN structure, the class attribute does not have any parents. Hence we have \begin{math}I(C; \Pi_C) = 0\end{math}. Also, each attribute has as its parents the class attribute and at most one other attribute. Hence for the ABN structure, we have
\begin{equation}
\sum_i^nI(X_i; \Pi_{X_i}) = \sum_{i, |\pi(i)|=2}^nI(X_i; \Pi_{X_i}, C) + \sum_{i, |\pi(i)|=1}^nI(X_i; C)
\end{equation}
The first term on R.H.S in equation (2) represents all attributes with an incoming augmenting arc. The second term represents attributes without an incoming augmenting arc.
Consider the chain law for mutual information given below
\begin{equation}
I(X; Y, Z) = I(X; Z) + I(X; Y|Z)
\end{equation}
Applying the chain law to the first term on R.H.S of equation (2) we get
\begin{equation}
\sum_i^nI(X_i; \Pi_{X_i}) = \sum_{i, |\pi(i)|=2}^nI(X_i; \Pi_{X_i}|C) + \sum_{i}^nI(X_i; C)
\end{equation}
For any ABN structure, the second term of equation (4) - \begin{math}\sum_{i}^nI(X_i; C)\end{math} is a constant. This is because, the term represents the arcs from the class attribute to all other attributes in the network, and these arcs are common to all ABN structures (as per the definition). Using equations (1) and (4), we rewrite the non-constant terms of the MDL score for ABN structures as follows
\begin{equation}
MDL(B_{Aug}) = \frac{|B_{Aug}|\log{N}}{2} - N\sum_{i, |\pi(i)|=2}^nI(X_i; \Pi_{X_i}|C)
\end{equation}
where, $B_{Aug}$ denotes an ABN structure.
\section{Some Insights}
Looking at the MDL score given in equation (5), we present a few insights on the learning ABN problem. The first term of the MDL equation - \begin{math}\frac{|B_{Aug}|\log{N}}{2}\end{math} represents the length of the ABN structure. Note that the length of any ABN structure depends only on the number of augmenting arcs, as the rest of the structure is the same for all ABNs. If we annotate the augmenting arcs with mutual information between the respective head and tail attributes, then the second term - \begin{math}N\sum_{i, |\pi(i)|=2}^nI(X_i; \Pi_{X_i}|C)\end{math} represents the sum of costs of all augmenting arcs. Since the best MDL score is the minimum score, our problem can be thought of as balancing the number of augmenting arcs against the sum of costs of all augmenting arcs, where we wish to maximize the total cost.

The MDL score for ABN structures is decomposable on attributes. We can rewrite equation (5) as 
\begin{equation}
\sum_i^n \left[\frac{|X_i|\log{N}}{2} - NI(X_i; \Pi_{X_i}|C)\right]
\end{equation}
where $|X_i|$ are the number of parameters stored at attribute $X_i$. The number of parameters stored at attribute $X_i$ depends on the number of parents of $X_i$ in $B_{Aug}$, and hence on whether $X_i$ has an incoming augmenting arc. Since we want to minimize the MDL score of our network, we should add an augmenting arc to an attribute $X_i$ only if its cost $I(X_j; X_i|C)$ dominates the increase in the number of parameters of $X_i$. For example, consider an attribute $X_i$ with no augmenting arc incident on it. Then the number of parameters stored at the attribute $X_i$ in ABN will be $||C||(||X_i|| - 1)$, where $||C||$ and $||X_i||$ are the number of states of the attributes $C$ and $X_i$ respectively. Thus $|X_i| = ||C||(||X_i|| - 1)$. If now an augmenting arc $e=(X_j, X_i)$ having a cost of $cost(e)=I(X_i; X_j|C)=I(X_j;X_i|C)$ is made incident on the attribute $X_i$, then the number of parameters stored at $X_i$ will be $|X_i| = ||X_j||.||C||.(||X_i|| - 1)$, where $||X_j||$ is the number of states of the attribute $X_j$. Note that the addition of the augmenting arc has increased the number of parameters of the network. Since we want to add an augmenting arc on $X_i$ only if it reduces the MDL score, the following condition must be satisfied
\begin{equation}
\frac{||C||(||X_i|| - 1)\log{N}}{2} > \frac{(||X_j||.||C||.(||X_i|| - 1))\log{N}}{2} - Ncost(e)
\end{equation}
which is equivalent to
\begin{equation}
cost(e) > \frac{||C||(||X_i|| - 1)(||X_j|| - 1)}{2N}\log{N} = T_R
\end{equation}
Note that this equivalence implies that the overall change in MDL score is independent of the arc direction. That is, adding an augmenting arc $(X_i, X_j)$ changes the network score identically to adding the arc $(X_j, X_i)$. Thus any augmenting arc is eligible to be added to an ABN structure if it has a cost at least the defined threshold $T_R$ and if it does not violate the ABN structure. Note that, this threshold depends only on the number of discrete states of the attributes and the number of cases in the input database, and is independent of the direction of the augmenting arc. We now present a polynomial time greedy algorithm for learning optimal ABN with respect to MDL score.
\section{The Algorithm}
\begin{enumerate}
\item Construct a complete undirected graph $G=(V,E)$, such that $V$ is the set of attributes (excluding the class attribute).
\item For each edge $e=(i, j) \in G$, compute $cost(e) = I(X_i; X_j|C)$. Annotate $e$ with $cost(e)$.
\item Remove from the graph $G$ any edges that have a cost less than the threshold $T_R$. This will possibly make the graph $G$ unconnected.
\item Run the Kruskal's Maximum Spanning Tree algorithm on each of the connected components of $G$. This will make $G$ a maximum cost forest (a collection of maximum cost spanning trees).
\item For each tree in $G$, choose a root attribute and set directions of all edges to be outward from the root attribute.
\item Add the class variable as a vertex $C$ to the set $V$ and add directed edges from $C$ to all other vertices in $G$.
\item Return $G$.
\end{enumerate}
The algorithm constructs an undirected graph $G$ in which all edges have costs above the defined threshold $T_R$. As seen in the previous section, all edges having costs greater than the threshold improve the overall score of the ABN structure. Running the Maximum Spanning Tree algorithm on each of the connected components of $G$ ensures that the ABN structure is preserved and at the same time maximizes the second term of the MDL score given in equation (5). Note that, if in step 3 of the algorithm the graph $G$ remains connected, our algorithm outputs a TAN structure. In this sense, our algorithm can be thought of as a generalization of the TAN algorithm given in \cite{friedman:bayesian}. The next section proves that the Augmented Bayes structure output by our algorithm is optimal with respect to the MDL score.
\section{Proof}
We prove that the ABN output by our algorithm is optimal by making the observation that no optimal ABN can contain any edge that was removed in step 3 of the algorithm. This is because, removing any such edge lowers the MDL score and leaves the structure an ABN. Consequently, an optimal ABN can contain only those edges that remain after step 3 of the algorithm. If an optimal ABN does not connect some connected component of the graph $G$ that results following step 3, edges with costs greater than or equal to $T_R$ can be added without increasing overall MDL score until the component is spanned. Hence there exists an optimal ABN that spans each component of the graph $G$ that results from step 3. By the correctness of Kruskal's algorithm run on each connected component to find a maximum cost spanning tree, an optimal ABN is found. Thus the ABN output by our algorithm is an optimal ABN.
\bibliographystyle{abbrv}

\end{document}